\pdfoutput=1

\documentclass[11pt]{article}

\usepackage{EACL2023}

\usepackage{times}
\usepackage{latexsym}
\usepackage{array} 
\usepackage{hyperref}

\usepackage[T1]{fontenc}

\usepackage[utf8]{inputenc}

\usepackage{microtype}

\usepackage{inconsolata}

\usepackage{multirow}
\usepackage{booktabs}
\usepackage{adjustbox}

\title{\texttt{WikiGoldSK}: Annotated Dataset, Baselines and Few-Shot Learning Experiments for Slovak Named Entity Recognition}

\author{D\'{a}vid \v{S}uba \quad Marek \v{S}uppa \quad Jozef Kub\'{i}k \quad Endre Hamerlik \quad Martin Tak\'{a}\v{c}\\
Comenius University in Bratislava, Slovakia\\
{\tt Contact: marek@suppa.sk}}

\begin{document}
\maketitle
\begin{abstract}
Named Entity Recognition (NER) is a fundamental NLP tasks with a wide range of practical applications. 
The performance of state-of-the-art NER methods depends on high quality manually anotated datasets which still do not exist for some languages. 
In this work we aim to remedy this situation in Slovak by introducing \texttt{WikiGoldSK}, the first sizable human labelled Slovak NER dataset. 
We benchmark it by evaluating state-of-the-art multilingual Pretrained Language Models and comparing it to the existing silver-standard Slovak NER dataset.
We also conduct few-shot experiments and show that training on a sliver-standard dataset yields better results.
To enable future work that can be based on Slovak NER, we release the dataset, code, as well as the trained models publicly under permissible licensing terms at \url{https://github.com/NaiveNeuron/WikiGoldSK}.
\end{abstract}

\section{Introduction}

Named Entity Recognition (NER) is a lower-level Natural Language Processing (NLP) task in which the aim is to both identify and classify named entity expressions in text into a pre-defined set of semantic types, such as Location, Organization or Person \cite{goyal2018recent}.
It is a key component of many downstream NLP tasks, ranging from information extraction, machine translation, question answering to entity linking and co-reference resolution, among others. 
Since its introduction at MUC-6 \cite{grishman1996message}, the task has been studied extensively, usually as a form of token classification. In recent years, the advent of pre-trained language models (PLMs) combined with the availability of sufficiently large high quality NER-annotated datasets has led to the introduction of NER systems with very high reported performance, sometimes nearing human annotation quality \cite{he2021debertav3}.

As the predominant method for adapting PLMs to a specific task of interest is model fine-tuning using training data, the availability of annotated NER datasets for both the training as well as the evaluation part of the process of creating a NER system is critical. 
Since their creation is expensive, many works have focused on extracting multilingual silver-standard NER datasets from publicly available corpora such as Wikipedia, exploiting the link structure to locate and classify named entities \cite{nothman2013learning, al2015polyglot, tsai2016cross, pan-etal-2017-cross}. 
While these methods have yielded NER-annotated datasets of significant size, with the recent follow-up work reporting quality comparable to that of datasets created via manual annotation \cite{tedeschi-etal-2021-wikineural-combined}, their application has multiple limitations: only a limited amount of Wikipedia text is inter-linked, mapping Wikipedia links to the pre-defined NER classes is non-trivial and their application often depends on the existence of high quality knowledge bases which may not be available for some domains and languages.

In this paper we focus on Slovak, a language of the Indo-European family, spoken by 5 million native speakers, which is still missing a manually annotated NER dataset of substantial size. To fill this gap, we propose the following contributions:

\begin{itemize}
    \item We introduce a novel, manually annotated NER dataset called \texttt{WikiGoldSK} built by annotating articles sampled from Slovak Wikipedia and labeled with four entity classes.
    \item We evaluate a selection of multilingual NER baseline models on the presented dataset to compare its quality with that of existing silver-standard Slovak NER datasets.
    \item We treat Slovak as a low-resource language and also assess the possibility of using few-shot learning to train a Slovak NER model using a small part of the introduced dataset.
\end{itemize}

\section{Related Work}
\paragraph{NER datasets}
Much of the progress in NER over the past decades can be attributed to and evidenced by the results reported on standard benchmarks, which in turn originate from shared tasks. This is because they generally provide high-quality annotation datasets, which are key both for the evaluation as well as creation of NER systems.
Shared tasks were first introduced for resource-rich languages, such as English, Spanish, German and Dutch \cite{tjong-kim-sang-2002-introduction,tjong-kim-sang-de-meulder-2003-introduction} and later established for other language groups, such as Indic \cite{Sangal:08} or Balto-Slavic languages \cite{piskorski-etal-2017-first,piskorski2019second,piskorski2021slav}. 
The ''First Cross-Lingual Challenge on Recognition, Normalization and Matching of Named Entities in Slavic Languages'' (BSNLP 2017) \cite{piskorski-etal-2017-first} is of particular relevance for our work, as to the best of our knowledge, it introduced the only publicly available human annotated Named Entity Recognition dataset based specifically on Slovak newswire. The dataset, however, consists of less than 50 human-annotated articles and can at best only be used for evaluation of Slovak NER systems but not their training.

The over-reliance on newswire text in the shared tasks has been noticed by the authors of \cite{balasuriya-etal-2009-named} who introduced the  manually annotated \texttt{WikiGold} dataset based on English Wikipedia articles. 
Despite its limited size, it is still used as an evaluation benchmark. 
As our aim is also to create a manually annotated (gold-standard) dataset based on Wikipedia articles, we use \texttt{WikiGoldSK} to refer to the dataset introduced in this work.

To alleviate the need for sizeable datasets at low cost across multiple languages, various methods of automatically generated NER-annotated datasets have been introduced.
In \cite{nothman2013learning} the authors introduce the WikiNER datasets, which makes use of Wikipedia articles and spans 9 languages but does not include Slovak.
Utilizing a similar approach, \cite{pan-etal-2017-cross} first classified English Wikipedia pages to specific entity types and then used the cross-lingual links to transfer the annotations to other languages. 
As not all entries are linked, the authors also utilize self-training and translation methods to match as many entries as possible.
This pipeline generates a dataset that covers 282 languages and includes Slovak as well.
With roughly 50 thousand entities annotated with categories Person, Location and Organization, it is the largest publicly available Slovak NER dataset to date.

Another approach of resolving the need for a sizable training dataset is to utilize few-shot learning, in which only a couple of expertly annotated training examples are provided.
Recently, the methods based on the combination of cloze-style rephrasing with language models have been shown to perform comparably to GPT-3 \cite{brown2020language} while having significantly fewer parameters \cite{schick2020s}. We consider a variant of Pattern-Exploiting Training~\cite{schick2020exploiting} called PETER \cite{la2021few} and to the best of our knowledge for the first time evaluate its performance for a general-purpose NER system in a specific language. 

\paragraph{Slovak NER}
The prior art in Slovak NER is limited.
In \cite{kavsvsak2012extrakcia} the authors identified potential named entities as words with capital letters and recognized new entities by finding the entity scope through Wikipedia parsing. For the purposes of this work they also created a dataset annotated by 60 human experts totalling 1620 entities. 
The authors of \cite{maruniak} and \cite{luptak} worked with datasets based on more than 5000 articles extracted from Slovak Wikipedia, containing more than 15 000 entities and used multiple well-established NLP toolkits and libraries (such as SpaCy) to train NER models on this dataset.
Utilizing a different datasource, \cite{mico} have focused on the Twitter account of one of the biggest Slovak journal and created a dataset with 10 000 of its NER-annotated tweets and almost 16 000 entities, and used it to train a NER model which utilized both FastText \cite{bojanowski2016enriching} vectors and the BiLSTM neural network architecture.
Unfortunately, none of the datasets and models mentioned in the aforementioned works are publicly available.

Despite having 5 million native speakers and being one of the official languages of the European Union, there are relatively few readily available NLP tools tailored specifically for Slovak, which might be to some extent caused by its linguistic and historical closeness to the much better resourced Czech. This creates a peculiar dichotomy: Slovak has too many native speakers to be considered ''low-resource'' but at the same time lacks readily available labelled datasets that are a prerequisite for many standard NLP tools. The ''language richness'' taxonomies such as \cite{joshi-etal-2020-state} consider Slovak among the ''The Rising Stars'' of languages, but it is, to the best of our knowledge, one of the few in this category that lacks a sizable, manually labelled NER dataset\footnote{The other languages lacking a sizable, manually NER-annotated datasets are Uzbek, Georgian, Belarusian, Egyptian Arabic and Cebauano.}. The introduction of SlovakBERT in \cite{pikuliak2021slovakbert} does suggest, however, that there is interest in creating Slovak-specific NLP tools and resources. Our work aims to help push this trend further.

\section{Dataset}

\begin{table}[]
    \centering
    \begin{tabular}{lrrr}
        \toprule
         & \textbf{WikiANN} & \textbf{BSNLP2017} & \textbf{\texttt{WikiGoldSK}} \\
         \toprule
         \# doc & N/A & 49 & 412\\
         \# sent & 30 000 & 741  & 6 696\\
         \# tok & 263 516 & 14 400& 128 944\\
         \midrule
         split & 2:1:1 & 0:0:1 & 7:1:2 \\
         \midrule
         \texttt{LOC} & 19 643 & 244 & 4 459\\
         \texttt{PER} & 18 238 & 255 & 2 739\\
         \texttt{ORG} & 15 286 & 273 & 1 929\\
         \texttt{MISC} &   N/A &  55 & 1 668\\
         \bottomrule
    \end{tabular}
    \caption{The comparison of \texttt{WikiGoldSK} to the other publicly available Slovak NER datasets. The terms \# doc, \# sent and \# tok refer to the number of documents, sentences and tokens in the specific datasets, respectively. Note that WikiANN does not provide document-level split and is not labeled with the \texttt{MISC} entity.}
    \label{tab:dataset_stats}
\end{table}

When creating the \texttt{WikiGoldSK} dataset, our principal aim was to create a high quality, publicly available human annotated corpora that could be used to both evaluate and build Slovak NER systems and that would be comparable to well-established benchmark datasets in other languages.
To ensure the resulting dataset can be used in the future for research as well as commercial use, we sampled 412 articles from the \texttt{skwiki-20210701} dump of Slovak Wikipedia, licensed under the terms of the Creative Commons Attribution-Share-Alike License 3.0.
In order for an article to be included, its last change date needed to be in 2021 and its length had to fit between 500 and 5 000 characters\footnote{This is motivated by the observation that long articles may shift the dataset towards their domain, whereas short articles often do not contain any named entity.}.
The raw text of the articles was tokenized by the generic English spaCy tokenizer, with a manual pass over the dataset in which the Slovak-specific tokenization mistakes were remedied.

\begin{table}[]
    \centering
    \begin{tabular}{lrrr}
        \toprule
         & \textbf{train} & \textbf{dev} & \textbf{test} \\
         \toprule
         \# sent & 4 687 & 669  & 1 340 \\
         \# tok & 90 616 & 12 794 & 25 534 \\
         split size & 70\% & 10\% & 20\% \\
         \midrule
         \texttt{LOC} & 3 040 & 461 & 958 \\
         \texttt{PER} & 1 892 & 298 & 549 \\
         \texttt{ORG} & 1 361 & 190 & 378 \\
         \texttt{MISC} & 1 184 & 160 & 324 \\
         \bottomrule
    \end{tabular}
    \caption{The frequency distribution across the \texttt{WikiGoldSK}'s train/dev/test splits.}
    \label{tab:dataset_splits}
\end{table}

We use the same set of tags as the CoNLL-2003 NER Shared task \cite{tjong-kim-sang-de-meulder-2003-introduction}, that is Location (\texttt{LOC}), Person (\texttt{PER}), Organization (\texttt{ORG}) and Miscellaneous (\texttt{MISC}), and our annotation guidelines are inspired by the ones introduced by the BSNLP 2017 shared task \cite{piskorski-etal-2017-first}.
The annotation was done using Prodigy\footnote{\url{https://prodi.gy/}} in which the whole dataset was pre-loaded with the labels predicted by the SlovakBERT model finetuned on the training part of the Slovak portion of the WikiANN dataset.
The dataset was annotated by three Slovak native speakers who are also authors of this paper.
Two annotators provided annotations for the full dataset whereas one annotator corrected half of the dataset.
The Cohen's kappa coefficient between the first two annotators is 0.90 when compared on the token level and 0.81 for the tokens where both annotators agreed that they were not a part of a named entity.
As per the benchmark established in \cite{landis1977measurement}, the coefficient values in both cases suggest ''almost perfect'' strength of agreement and a high quality of the annotation.
To arrive at the final dataset, the ambiguities were resolved in a discussion between the annotators.

The summary statistics of the resulting dataset, along with the existing Slovak NER datasets, can be found in \autoref{tab:dataset_stats}. 
As we can see, \texttt{WikiGoldSK} is larger than the Slovak portion of the BSNLP2017 dataset but smaller than the Slovak portion of the WikiANN dataset.
At the same time, one can see that the distribution of Named Entities in WikiANN and \texttt{WikiGoldSK} follows the same pattern, with the order of \texttt{LOC}, \texttt{PER}, \texttt{ORG} holding for both datasets in terms of entity frequency, which is not the case in BSNLP2017.
This is probably caused by the fact that both  WikiANN and \texttt{WikiGoldSK} are based on Wikipedia articles whereas BSNLP2017 is based on newswire text.

To make the dataset compatible with existing benchmarks, we also introduce a standard train/dev/test split in the 7:1:2 ratio, described in detail in \autoref{tab:dataset_splits}. 
We note that the size of the test portion of \texttt{WikiGoldSK} is on the same order as that of \texttt{WikiGold} which consists of 1 696 sentences and 39 007 tokens. 

\section{Experiments}

We conduct two types of experiments with the newly introduced dataset. 
First, we establish a set of baselines based on existing state-of-the-art PLMs that were pre-trained on Slovak data. 
Next, we emulate a low-resource setup by only using a small sample of the training set and use it to evaluate a few-shot learning approach as well.

\subsection{Baselines}

To evaluate a broad set of baselines on \texttt{WikiGoldSK}, we choose three well-established NLP toolkits:
\begin{itemize}
    \item \textbf{spaCy\footnote{\url{https://spacy.io}}}, which provides a pipeline for converting words to embedding of user's choice and then models NER as a structured prediction task,
    \item \textbf{Trankit}~\cite{trankit}, which is based on XLM-RoBERTa \cite{xlm-roberta}, provides pre-trained models for 56 languages, including Slovak, along with the ability to finetune on custom NER datasets, and
    \item \textbf{Transformers}~\cite{huggingface}, which has become the standard tool for training, storing and sharing Transformer-based models and also includes readily available scripts for finetuning PLMs on NER datasets.
\end{itemize}

When it comes to the models chosen as baselines, we again chose well-established models relevant to the task of Slovak NER:

\begin{itemize}
    \item \textbf{XLM-RoBERTa} (\texttt{XLM-R-base}), a multilingual Transformer model pretrained on text spanning 100 languages, including Slovak,
    \item \textbf{SlovakBERT}, the only BERT-based model specifically optimized for Slovak, which was pre-trained on almost 20GB of Slovak text obtained from various sources, including crawling Slovak web, and
    \item \textbf{mDeBERTav3}~\cite{he2021debertav3}, a multilingual Transformer model pretrained on the same dataset as XLM-RoBERTa using a different training objective which leads to more efficient training and better performance on various benchmarks.
\end{itemize}

Our experiments were generally conducted by finetuning a given model using a specific NLP toolkit on a selected dataset, while utilizing the test set of the \texttt{WikiGoldSK} for evaluation. We use three datasets for finetuning: WikiANN, WikiANN combined with  \texttt{WikiGoldSK} and just \texttt{WikiGoldSK}. Only the training portions of the respective datasets were used for finetuning. Additionally, we also benchmark the models trained on the \texttt{WikiGoldSK} dataset on the Slovak portion of the BSNLP2017 dataset.

\subsection{Few-shot learning}

To evaluate the possibility of building a Slovak NER system, we chose the PETER (PET \cite{schick2020exploiting} for NER) method introduced in \cite{la2021few}. 
At its core, it uses pattern-verbalizer pairs (PVP), in which the ''pattern'' part converts a sentence with a token that corresponds to a named entity and creates a cloze-style phrase containing exactly one \texttt{[MASK]} token and the ''verbalizer'' maps tokens predicted by a PLM in place of \texttt{[MASK]} to one of the considered Named Entity classes.
Each labeled sentence $s$ is converted into $|s|$ pairs of training inputs $x = (s, t)$ where $t$ is a particular token from the sentence we are predicting a label to; the training set then consist of pairs $(x, y)$ where $y$ is the ground-truth label.
A separate language model $M$ is fine-tuned for each PVP, a soft-label dataset created from unlabeled data and finally, the resulting classifier is trained on this dataset.

In our experiments, we use two PVPs below. More details can be found in \autoref{sec:appendix_peter}.

\begin{itemize}
    \item $P_1((s, t))$: "$s$. V predchádzajúcej vete slovo $t$ označuje entitu \texttt{[MASK]}."~(English translation: \emph{''$s$. In the previous sentence, the word $t$ refers to a/an \texttt{[MASK]} entity.})
    \item $P_2((s, t))$: "$s$. $t$ je \texttt{[MASK]}."~(English translation: \emph{''$s$. $t$ is a \texttt{[MASK]}.})
\end{itemize}

\begin{table*}[h]
\centering
\begin{adjustbox}{width=1\textwidth}
\begin{tabular}{lcccccccccccc}
\toprule
\multicolumn{1}{c}{\multirow{2}{*}{}}                                      & \multicolumn{3}{c}{\textbf{WikiANN}}                                                      & \multicolumn{3}{c}{\textbf{WikiANN} + \textbf{\texttt{WikiGoldSK}}}                                         & \multicolumn{3}{c}{\textbf{\texttt{WikiGoldSK}}}   & \multicolumn{3}{c}{\textbf{BSNLP2017}}                                                \\
\multicolumn{1}{c}{}                                                       & P                         & R                         & F1                        & P                         & R                         & F1                        & P                         & R                         & F1    & P                         & R                         & F1                     \\ \toprule
\textbf{spaCy}                                                               &                           &                           &                           &                           &                           &                           &                           &                           &                           \\
XLM-RoBERTa                                                                  & 0.5639                           & 0.7413                          & 0.6405                          & 0.8809                          & 0.8973                           & 0.8890                          & 0.9145                           & 0.8955                           & 0.9049   & \textbf{0.8102}                          & 0.7722                          & 0.7907                          \\
SlovakBERT                                                                   & 0.5509                           & 0.7285                           & 0.6274                           & 0.8754                           & 0.8932                          & 0.8842                          & 0.8889                           & 0.9122                           & 0.9004     & 0.7186                          & 0.7704                          & 0.7436                          \\
mDeBERTaV3                                                              & 0.5925                           & \textbf{0.7572}                          & \textbf{0.6648}                          & 0.8621                          & 0.8855                           & 0.8737                          & 0.9151                          & 0.9167                          & 0.9159   & 0.8024                          & 0.8122                          & 0.8073                          \\
\midrule
\textbf{Trankit}                                                             &                           &                           &                           &                           &                           &                           &                           &                           &                           \\
XLM-RoBERTa                                                                  & \textbf{0.6110}                           & 0.7020                          &  0.6533                          & 0.8833                          & 0.8805                          & 0.8819                          & 0.8869                          & 0.9014                          & 0.8941 & 0.7882                           &  0.8252                         & 0.8063                          \\ \midrule
\textbf{Transformers}                                                        &                            &                           &                            &                           &                           &                           &                           &                           &                           \\
XLM-RoBERTa                                                                  & 0.5247                          & 0.7423                          & 0.6148                          & 0.8815                          & 0.9018                           & 0.8915                          & 0.9210                           & 0.9339                           & 0.9274    & 0.7760                           &  0.8226                         & 0.7986                          \\
SlovakBERT                                                                  & 0.5265                           & 0.7428                           & 0.6162                           & \textbf{0.9020}                           & \textbf{0.9208}                          & \textbf{0.9113}                          & 0.9179                           & 0.9262                           & 0.9221   & 0.7900                           &  0.8278                         & \textbf{0.8085}                           \\
mDeBERTaV3                                                                  & 0.5092                           & 0.7471                            & 0.6056                           &  0.8835                         & 0.9063                           & 0.8948                          & \textbf{0.9302}                           & \textbf{0.9412}                           & \textbf{0.9357}   & 0.7793                           &  \textbf{0.8322}                         & 0.8049                            \\
\bottomrule
\end{tabular}
\end{adjustbox}
\caption{The results of finetuning various baselines using the three selected NLP toolkits on three dataset combinations and evaluating on the test set of \texttt{WikiGoldSK}. The P, R and F1 refer to Precision, Recall and the F1 score, respectively. Best result per metric and dataset is boldfaced.}
\label{tab:baselines_results}
\end{table*}

\section{Results}

The results of the evaluation of baselines can be seen in \autoref{tab:baselines_results}. 
They suggest that XLM-RoBERTa can still be considered a strong baseline, as its performance is similar to that of SlovakBERT, despite the latter being specifically trained and optimized for Slovak.
Across the three NLP toolkits, we observe that the performance of Trankit is generally lower than that of spaCy and Transformers, given the same dataset.
Comparing the three models finetuned either using spaCy or Transformers, \autoref{tab:baselines_results} suggests that mDeBERTaV3 obtains performance that is either very similar or better than that of XLM-RoBERTa across all considered datasets.
A model based on mDeBERTaV3 also reported the best performance out of all models evaluated on \texttt{WikiGoldSK} and performance on par with SlovakBERT on the BSNLP2017 dataset. 
Finally, we also note that the choice of the training dataset has significant impact on the performance of the resulting NER model, as the difference between the F1 scores of the best performing model on the WikiANN dataset and the \texttt{WikiGoldSK} dataset is over 0.27.
Despite the much larger size of the WikiANN dataset, the results in \autoref{tab:baselines_results} suggest it is best not to combine it with the manually annotated dataset in order to obtain the best results.

\begin{table}[h]
\centering
\begin{tabular}{lccccccccc}
\toprule
\multicolumn{1}{c}{\multirow{2}{*}{}} & P                         & R                         & F1               \\ 
\toprule
\textbf{10 shots}   &           &             &          \\
PVP 1  & 0.4262         & 0.5290              & 0.4720    \\
PVP 2  & 0.4320         & 0.6163              & 0.5079    \\
PVP 1 \& 2  & 0.4834          & 0.5937              & 0.5329     \\
\textbf{30 shots}   &           &             &          \\
PVP 1  & 0.4853         & 0.5968              & 0.5353    \\
PVP 2  & 0.4921         & 0.6502              & 0.5602    \\
PVP 1 \& 2  & 0.4857          & 0.6072              & 0.5397    \\
\textbf{50 shots}   &           &             &          \\
PVP 1  & 0.5198          & 0.6176              & 0.5645     \\
PVP 2  & 0.5041         & \textbf{0.6688}              & 0.5749    \\
PVP 1 \& 2  & \textbf{0.5321}         & 0.6484           & \textbf{0.5845}     \\
\bottomrule
\end{tabular}
\caption{The results of the PETER few-shot experiments for various shots and combinations of patter-verbalizer pairs (PVP) in terms of Precision (P), Recall (R) and F1 score. Best results are boldfaced.}
\label{tab:peter_results}
\end{table}

When it comes to the few-shot learning experiments, the results can be seen in \autoref{tab:peter_results}.
We note that the combination of PVP 1 and PVP 2 (denoted "PVP 1 \& 2" in \autoref{tab:peter_results}) yields better results than when they are used separately.
Comparing the results with those presented in \autoref{tab:baselines_results}, we can see that the supervised models outperform the few-shot learning approaches, even when trained on a silver-standard dataset.
This suggests that more work is necessary for few-shot NER approaches to be competitive with supervised approaches.

\section{Conclusion}

In this work, we introduce \texttt{WikiGoldSK}, the first sizable, manually annotated NER dataset in Slovak. We have established first baseline benchmarks on the dataset using state-of-the-art models, including multilingual and Slovak-specific models. The experiments with few-shot learning suggest that its performance does not reach that of supervised learning. The \texttt{WikiGoldSK} dataset is publicly released under permissible licensing terms, enabling training and evaluation of future models as well as tracking the progress in Slovak NER.

\section*{Limitations}

While \texttt{WikiGoldSK} is currently the largest manually annotated Slovak NER dataset, it is still small in the great scheme of things, especially when its size (roughly 10 thousand labelled entities) gets compared to that of the CoNLL-2003 or Czech Named Entity Corpus 2.0 datasets (both with 35 thousand labelled entities). Our few-shot experiments have only been conducted in the case of Slovak and the newly introduced dataset, and may not generalize to other languages and datasets.

\section*{Ethics Statement}
The dataset used for annotation was sampled from Slovak Wikipedia, which allows for reuse of its content under the terms of the Creative Commons Attribution-Share-Alike License 3.0. The annotated dataset is released under the same license.

\section*{Acknowledgements}

This project was supported by grant APVV-21-0114.

\bibliography{anthology,custom}
\bibliographystyle{acl_natbib}

\appendix

\newpage

\section{Annotation Manual}
\label{sec:appendix_annotation_manual}

For the purpose of the \texttt{WikiGoldSK} dataset, we define the following classes of Named Entities:

\begin{itemize}
    \item \textbf{PER} Names, surnames, nicknames of living beings, without titles. Groups of people that belong to a nation, city, family..., e.g. \texttt{Slováci} (\emph{Slovaks} in English), \texttt{Bratislavčania} (meaning: people who live in the city of Bratislava), \texttt{Kováčovci} (meaning: family name). General adjectives are \textbf{not} entities e.g. \texttt{rímsky vojak} (\emph{roman soldier} in English), \texttt{slovenský jazyk} (\emph{slovak language} in English), but personal adjectives \textbf{are} PER entity, e.g. in \texttt{"to je Petrov kufor"} (\emph{"it's Peter's suitcase"} in English), \texttt{"Petrov"} is a PER entity.
    
    \item \textbf{LOC} All territorial and and geo-political units, such as countries, cities, regions... Physical locations as rivers, parks, buildings, bridges, castles, roads... Streets were also classified as LOC entities, but without building numbers.
    
    \item \textbf{ORG} Political parties, companies, government institutions, political/sport/educational organizations, music bands. Museums, zoos and theaters were also annotated as ORG, although they are very close to LOC. However, in our opinion, their meaning exceed the location aspect. But if the context makes it clear that the described object is only a building and/or an area that belongs to an organisation, LOC should be used. Companies were also labeled with legal suffix, e.g. \texttt{ESET, spol. s r.o.} is all standing for one ORG entity. 
    
    \item \textbf{MISC} Names of movies, awards, events, festivals, newspapers, TV or radio station names. Also sport series, cups and leagues were annotated as MISC.
\end{itemize}
In case of nested entities, the outer one is recognized as entity, e.g. whole \texttt{"Národná Banka Slovenska"} (\emph{National Bank of Slovakia} in English) is ORG entity. Abbreviations following entity is separate entity, e.g. in \texttt{"Úrad verejného zdravotníctva (UVZ)"} (\emph{Office of Public Health (OPH)} in English) we annotate 2 separate ORG entities.

The main differences between our guidelines and that of the BSNLP 2017 shared task \cite{piskorski-etal-2017-first} are as follows:
\begin{itemize}
    \item For entities such as museums, theathers, zoos we've preferred ORG entity and only if it's clear from the context, LOC could be used. However, in BSNLP 2017 shared task these entities were always annotated as LOC.
    \item We've used MISC entity for newspapers, TV or radio stations. In BSNLP 2017 shared task guidelines it's not explicitly stated but in dataset these entities are mostly annotated as ORG.
\end{itemize}

\section{PETER training details}
\label{sec:appendix_peter}

The unlabeled dataset is created by sampling 1000 sentences from the train split of \texttt{WikiGoldSK}. As a base model for training, we used SlovakBERT. To make the prediction comparable with that of the baselines, the token-level predictions were converted to the IOB2 form using a simple heuristic: whenever there is a sequence of entities of the same type, the tag of the very first entity is prefixed with \texttt{B-} while the rest is prefixed with \texttt{I-}. Note that this is a very imperfect heuristic, as it for instance cannot handle the cases where two entities from the same class are following straight after each other.

\end{document}